\documentclass[10pt,journal,compsoc]{IEEEtran}
\usepackage{amsmath,amsfonts}
\usepackage{algorithmic}
\usepackage{algorithm}
\usepackage{array}
\usepackage[caption=false,font=normalsize,labelfont=sf,textfont=sf]{subfig}
\usepackage{textcomp}
\usepackage{stfloats}
\usepackage{url}
\usepackage{verbatim}
\usepackage{graphicx}
\usepackage{cite}
\usepackage{ragged2e}
\usepackage{mathrsfs}
\usepackage{multirow}
\usepackage{color}
\usepackage{tikz}
\usepackage{booktabs}
\usepackage{threeparttable}

\begin{document}

\title{Bio-inspired Color Constancy: From Gray Anchoring Theory to Gray Pixel Methods} %

\author{Kai-Fu Yang,~\IEEEmembership{Member,~IEEE}, Fu-Ya Luo, and Yong-Jie Li,~\IEEEmembership{Senior Member,~IEEE}
\IEEEcompsocitemizethanks{\IEEEcompsocthanksitem Kai-Fu Yang and Yong-Jie Li are with the MOE Key Laboratory for Neuroinformation, School of Life Science and Technology, University of Electronic Science and Technology of China, Chengdu 610054, China. (Corresponding author: Kai-Fu Yang and Yong-Jie Li; Email: yangkf@uestc.edu.cn, liyj@uestc.edu.cn)}
\IEEEcompsocitemizethanks{\IEEEcompsocthanksitem Fu-Ya Luo is with the School of Electronic Engineering and Automation, Guilin University of Electronic Technology, Guilin, Guangxi 541004, China. Email: luofuya@guet.edu.cn}}

\markboth{Yang, et al.}%
{Shell \MakeLowercase{\textit{et al.}}: A Sample Article Using IEEEtran.cls for IEEE Journals}


\IEEEcompsoctitleabstractindextext{
\begin{abstract}
\justifying
Color constancy is a fundamental ability of many biological visual systems and a crucial step in computer imaging systems. Bio-inspired modeling offers a promising way to elucidate the computational principles underlying color constancy and to develop efficient computational methods. However, bio-inspired methods for color constancy remain underexplored and lack a comprehensive analysis. This paper presents a comprehensive technical framework that integrates biological mechanisms, computational theory, and algorithmic implementation for bio-inspired color constancy. Specifically, we systematically revisit the computational theory of biological color constancy, which shows that illuminant estimation can be reduced to the task of gray-anchor (pixel or surface) detection in early vision. Subsequently, typical gray-pixel detection methods, including Gray-Pixel and Grayness-Index, are reinterpreted within a unified theoretical framework with Lambertian reflection model and biological color-opponent mechanisms. Finally, we propose a simple learning-based method that couples reflection-model constraints with feature learning to explore the potential of bio-inspired color constancy based on gray-pixel detection. Extensive experiments confirm the effectiveness of gray-pixel detection for color constancy and demonstrate the potential of bio-inspired methods.

\end{abstract}

}

\maketitle

\section{Introduction}
\IEEEPARstart{C}{olor} constancy is a fundamental ability of many biological visual systems and a crucial step in computer imaging systems \cite{foster2011color, gijsenij2011computational}. In addition, efficiently removing the color cast triggered by light source is a prerequisite for ensuring that the extracted features of the scenes remain constant under varying illumination conditions \cite{xue2021does}. However, estimating the illuminant from a color-biased scene is challenging the problem is inherently ill-posed \cite{foster2011color, gijsenij2011computational}. Therefore, color constancy remains a fundamental perceptual problem that attracts sustained attention from both neuroscientists and computer vision researchers.

In the past few decades, most color constancy methods have been considered statistic- or physics-based methods \cite{gijsenij2011computational}, which are usually based on specific assumptions for the spatial and/or spectral characteristics of scenes. Recent advances have been significantly driven by machine learning, especially deep-learning techniques \cite{hu2017fc4,yu2020cascading, lo2021clcc,afifi2021cross}. However, because color constancy is a well-known color perception phenomenon, we believe that investigating the computational principles of color constancy in the biological visual system and developing bio-inspired methods is still a promising way for low-level computer vision. As an important intersectional issue of biological vision and computer vision, bio-inspired color constancy has not been systematically explored due to a limited understanding of visual mechanisms and computational theory. This fact motivates us to conduct a comprehensive investigation, spanning biological mechanisms, computational theory, and algorithmic implementation for bio-inspired color constancy.

The first issue is identifying potential computational principles for achieving color constancy in biological visual systems. Neuroscience and cognitive research have already brought some impressive insights \cite{land1971lightness, hurlbert2004color, dorr1996goldfish, werner1988color}. For instance, Land and McCann’s Retinex theory posits that the visual system recovers surface reflectance by computing lightness ratios between regions and anchoring the brightest surface to white \cite{land1971lightness}, which also inspires the well-known White-Patch method for color constancy. Meanwhile, color-opponent receptive fields in primary visual cortex (V1) have been proposed as the functional substrate for color constancy \cite{gegenfurtner2003cortical, hurlbert2003colour, shapley2011color} and have been partly verified in computational models \cite{gao2013color, gao2015color}. These studies fail to provide a unified computational framework of color constancy in the visual system. Recently, our theoretical analysis suggested a new gray-anchoring theory for color constancy in early vision \cite{yang2026gray}, which bridges gaps among color opponency mechanisms, Retinex theory, and illuminant estimation. Therefore, it offers an opportunity to establish a unified, systematic computational route for bio-inspired color constancy technologies and directly motivate this work.

From a computational perspective, illuminant estimation is equivalent to gray pixel detection. Once a gray pixel is identified, perfect color constancy can be achieved under uniform illumination, since the reflectance values of gray pixels are equal across the RGB channels. As suggested by \cite{yang2026gray}, illuminant estimation could be reduced to the task of gray anchor detection in early vision. Meanwhile, color-opponent mechanisms throughout early vision provide the computational basis of neural implementation for gray anchoring. Therefore, the well-known gray pixel detection methods, for example Gray-Pixel \cite{yang2015efficient} and Grayness-Index \cite{qian2019finding}, could be considered specific algorithm implementations of gray anchoring theory \cite{yang2026gray}, and hence are highly linked to color constancy in the early vision. Moreover, gray pixel detection has been widely studied in the field of computational color constancy and shown promising potential for illuminant estimation \cite{yang2015efficient,qian2019finding}. In the context of this study, we rethink typical gray pixel detection frameworks under the biological viewpoint and try to provide a unified theoretical description under the Lambertian reflection model and biological color-opponent mechanisms.

Finally, considering the biological visual system is still only partially understood, bio-inspired methods built on incomplete mechanisms may fail to reach optimal performance. For example, higher-level cortices are considered to provide prior for color constancy, but the computational principles remain unclear \cite{olkkonen2008color, hansen2006color}. Therefore, at the present stage, coupling biological insights with deep-learning-based feature learning is a reasonable trade-off which offers a practical route to promote bio-inspired methods to perform robustly in real-world scenes. For instance, our previous successful experience on integrating the visual light adaptation rules into a learnable network for image enhancement \cite{yang2023learning}. This encourages us to build a model-constrained feature learning method to further explore the potential of bio-inspired color constancy based on gray pixel detection.

In summary, we aim to systemically investigate and build a comprehensive technical route of the bio-inspired color constancy and offer new technical insights for color perception in low-level computer vision. To address these issues, the main contributions of this work are summarized as follows:

\begin{itemize}
\item This study presents the first attempt to establish a complete technical route for bio-inspired color constancy by integrating biological mechanisms, computational theories, and algorithmic implementations, thereby bridging the gap between neuroscience insights and computer vision applications.

\item In the context of bio-inspired color constancy, this work theoretically unifies the Gray-Pixel \cite{yang2015efficient} and Grayness-Index \cite{qian2019finding} methods under the Lambertian reflection model, at the same time, aligning both in the gray-anchoring theory of biological color constancy.

\item To verify the potential of bio-inspired gray pixel detection for color constancy, we propose a new model-constrained gray pixel detection network (GPNet) that embeds biological constraints into a feature learning network for robust illuminant estimation. Experimental results demonstrate that the proposed GPNet achieves comparable performance with state-of-the-art deep learning methods. 

\end{itemize}

\section{Related Work}
\subsection{Statistic- or Physics-based Methods} 
Statistic-based methods usually employ the simple Lambertian reflection model in which the specular reflection is ignored. The classical Gray-World hypothesis suggests that, under white illumination, the spatial average of surface reflectance in a scene is achromatic \cite{buchsbaum1980spatial, hurlbert1986formal}; consequently, the illuminant can be estimated directly from the average values of image channels. On the other hand, the White-Patch assumption, originating from the Retinex theory \cite{land1971lightness}, treats the highest value in each color channel as a perfect reflector; all other surfaces are then recovered by anchoring this brightest surface to white. Gray-Edge methods exploit higher-order image statistics and posit that the average spatial derivative in each color channel is achromatic \cite{van2007edge}. These methods and their generalized variants, such as Shades-of-Gray \cite{finlayson2004shades}, have been widely used in practice, considering the merits of easy implementation and low computational cost.

Unlike the Lambertian assumption, the dichromatic reflection model captures the physically more realistic light–surface interaction. Therefore, there are some methods that leverage knowledge of the physical interactions with the dichromatic reflection model \cite{finlayson2001solving, toro2008dichromatic}. For example, various methods estimate the illuminant using specularities or highlights \cite{lee1986method, healey1991estimating, tan2004color}. More literatures on these statistic- or physics-based methods for computational color constancy can be found in the impressive review by Gijsenij et al. \cite{gijsenij2011computational}.

\subsection{Learning-based Methods} 
Classical machine learning techniques are used to combine the outputs of multiple methods and to provide a universal illuminant estimation \cite{cardei1999committee,bianco2008consensus,schaefer2005combined, finlayson2013corrected}. In contrast, others are mainly based on middle- or high-level semantic information to decide which algorithm is likely to work better for a specific scene category (e.g., indoor or outdoor) \cite{bianco2010automatic, bianco2014adaptive, van2007using, gijsenij2011color}. Specifically, gamut-based methods assume one observes only a limited number of colors for a given illuminant in real-world images \cite{forsyth1990novel,barnard2000improvements, gijsenij2010generalized}. Other regression-based algorithms aim to directly map the imagery representation to a illumination vector \cite{funt2004estimating, cheng2015effective,qian2016deep}.

Recently, great progress has been made in color constancy in the deep-learning era, associated with the success of convolutional neural networks (CNN) on various vision tasks. First attempts of illuminant estimation with deep convolutional neural networks have been made and obtained promising performance \cite{bianco2015color, lou2015color, shi2016deep, bianco2017single}. Hu et al. introduce a fully convolutional framework that assigns spatially-varying confidence weights to image patches for improving illuminant estimation \cite{hu2017fc4}. In addition, Yu et al. improve robustness of illuminant estimation across datasets by cascading multiple CNN modes \cite{yu2020cascading}. Specifically, Barron et al. introduced Convolutional Color Constancy (CCC) \cite{barron2015convolutional} and Fast Fourier Color Constancy (FFCC) \cite{barron2017fast}, which formulate color constancy as a 2D spatial localization task within a log-chroma histogram space and achieve state-of-the-art performance on standard benchmarks. Recent advances include Integral Fast Fourier Color Constancy for multi-illuminant scenes\cite{wei2025integral} and a new method for cross-camera color constancy \cite{afifi2021cross}.

\subsection{Bio-inspired Methods} 
Historically, bio-inspired color constancy methods can be traced back to the Land's retinex theory \cite{land1964retinex,land1971lightness}. Retinex and its variants occupy important positions in computational color constancy methods. Besides, bio-inspired color constancy methods are developed to emulate various biological mechanisms across multiple stages of the human visual system. For example, Zhang et al. proposed a color constancy method by imitating the information processing mechanisms of the inner network of retina \cite{zhang2016retinal}. Meanwhile, Gao et al. developed a full framework of color-opponent mechanisms and preliminarily verified the core roles of V1 double-opponent cells in color constancy \cite{gao2013color, gao2015color}. In addition, Akbarinia and Parraga presented a bio-inspired color constancy model based on the adaptive surround modulation of cortical neurons \cite{akbarinia2017colour}. In this model, illuminant cues are obtained from the receptive field outputs with dynamic properties based on scene contrast. Recently, Ulucan et al. built a bio-inspired system that achieves color constancy and shows potential in explaining color assimilation illusions \cite{ulucan2022bio}.

However, current bio-inspired methods mainly leverage isolated biological principles, and a comprehensive computational framework of the bio-inspired color constancy has yet to be established. Our recent theoretical analysis of color constancy in early vision \cite{yang2026gray} offers an opportunity to bridge gaps between biological mechanisms and computational color constancy. Therefore, the main ambition of this study is to establish a complete technical route for bio-inspired color constancy that formalizes the transition from biological theory to computational practice.

\section{Color Constancy Theory in Early Vision}
\subsection{Revisiting Gray-Anchoring Theory}
\label{sec.ga}

\begin{figure*}[t!]
\centering
\includegraphics[width=18cm]{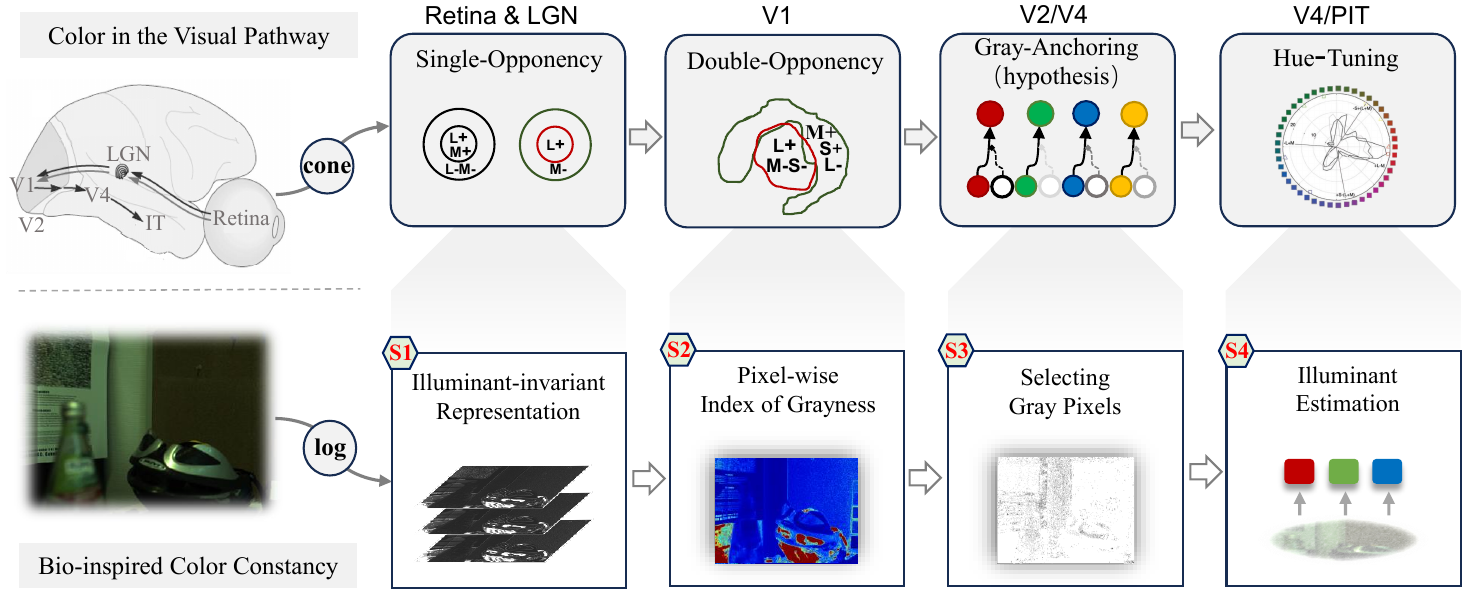}
\caption{The general framework of bio-inspired color constancy according to the gray anchoring theory. The input signals are first transformed into logarithmic space to facilitate the reflectance recovery or illuminant removal. Single- and Double-Opponent operators are subsequently implemented in the retina-to-V1 pathway for identifying the gray anchors (pixels or surfaces). Finally, gray anchors are selected in the V2 or V4 by combining the response of double-opponent cells, and the illuminant can be estimated (perhaps in V4 or posterior inferotemporal cortex (PIT)) via anchoring to these gray anchors.}
\label{FigGAflow}
\end{figure*}

To systematically organize the technical route of bio-inspired color constancy, we first simply review the computational mechanisms of color constancy in the biological visual system. Generally speaking, neuroscientists and cognitive scientists believe that multiple neural mechanisms in the visual system operate color constancy at different levels \cite{foster2011color}. Experimental data support that cone-excitation ratios and cone adaptation taking place in the retina contributes partly to color constancy \cite{foster1994relational, lee1999horizontal}. In the primary visual cortex, Conway and Livingstone found the double-opponent cells whose spatially and chromatically opponent receptive fields are critical for color constancy \cite{conway2001spatial, conway2006spatial}, indicating that V1 and lower visual areas may contribute more significantly to color constancy than previously estimated \cite{hurlbert2004color}. In addition, the role of V4 for color constancy was investigated by Zeki \cite{zeki1983colour}, Wild et al. \cite{wild1985primate} and Walsh et al. \cite{walsh1993effects}. Beyond the visual cortices, feedback from the higher cognitive tasks is also believed to contribute to color constancy, such as memory color effect \cite{witzel2018color} and familiar object recognition \cite{olkkonen2008color}, especially in challenging scenes.

However, these investigations on neural mechanisms have not provided the unified computational principles for biological color constancy. To address this issue, our recent work has proposed a gray-anchoring theory to explain how the color-opponent mechanisms in the early vision contribute to color constancy \cite{yang2026gray} (Figure \ref{FigGAflow}(top)). It also demonstrates that identifying gray surfaces within complex scenes is an important prerequisite task for biological color constancy \cite{yang2026gray}. According to gray-anchoring theory, logarithmic operator, single-opponency, and double-opponency are core steps for detecting gray anchors (pixels or surfaces) from the natural scenes. These steps can be reasonably implemented by the cells from retina to V2, with widely-studied color-opponent receptive fields. Then, the higher-level cortices (e.g., V4 or IT) will finish the illuminant estimation based on the efficient illuminant cues from the gray anchors. Our gray-anchoring theory provides a unified computational strategy for biological color constancy and systemically reveals the role of concentric double-opponent cells in gray anchor detection.

Computationally, existing color constancy methods, including statistic-based methods \cite{joze2012role}, bio-inspired methods \cite{ulucan2022bio}, and deep-learning-based methods \cite{bianco2019quasi}, have demonstrated a bias toward gray or white information in the illuminant estimation. In addition, earlier bio-inspired color constancy methods have directly investigated the role of V1 double-opponent cells in illuminant estimation \cite{gao2013color, gao2015color}, but they failed to clarify the computational principle of color constancy because they omitted some essential operators, like the logarithmic transform for the visual inputs \cite{marr2010vision, fechner1966elements, land1971lightness}. Mathematically, the logarithmic operator serves as a computational prerequisite for reflectance recovery, as it translates arithmetic division into a biologically plausible subtractive inhibition mechanism \cite{land1971lightness}. Meanwhile, the functional role of the logarithmic transform is also suggested by D. Marr when discussing the potential functional role of double-opponent cells in V1 \cite{marr2010vision}.

Recently, Ulucan et al. suggested a similar hierarchical framework for explaining color constancy and color assimilation illusions \cite{ulucan2022bio}. This framework also ignores the logarithmic transform for input signals, but shares a similar salient pixel statistics principle for illuminant estimation based on the double-opponent response. In addition, the double-opponent responses are formed by taking the absolute maximum response of single-opponent cells, which lacks sufficient physiological support. In contrast, the gray-anchoring theory provides a unified description that integrates these recognized computational principles and has a solid basis in neural computation.

In summary, we propose the general framework of bio-inspired color constancy for illuminant estimation by identifying and anchoring to gray pixels. Therefore, the illuminant estimation or color constancy task reduces to gray anchor detection in the proposed bio-inspired framework. Meanwhile, these computational steps of gray anchor detection are strongly grounded in the physiological evidence for color opponent receptive fields in visual pathway \cite{yang2026gray}. Finally, Figure \ref{FigGAflow}(bottom) shows the general computational flowchart of bio-inspired color constancy based on the gray anchoring theory. Computationally, the general steps of bio-inspired color constancy can be summarized as: (S1) obtaining the illuminant-invariant measures with the single-opponent operator on the logarithmic image, (S2) computing the index for gray of pixels via concentric double-opponent receptive fields, (S3) selecting gray anchors according to index for gray, and (S4) estimating illuminant by summing the hue of gray pixels.

\subsection{Rethinking Gray Pixel Detection}
\label{sec.gp}
From the view of computational color constancy, finding gray anchors from visual scenes is also a very attractive issue due to the fact that achromatic pixels or surfaces intrinsically encode illuminant information \cite{xiong2007automatic,li2010color, ono2022degree}. Specifically, the Gray-Pixel method \cite{yang2015efficient} and its improved variants \cite{qian2019finding, cheng2024nighttime} have systematically shown that illuminant estimation or color constancy can be effectively achieved via detecting gray pixels from color-biased images. According to the description of bio-inspired color constancy and gray-anchoring theory in section \ref{sec.ga}, gray-pixel-based methods actually can be considered specific implementations of the gray-anchoring rule, but how they are linked to bio-inspired color constancy remains incomprehensible.

In this section, we rethink the gray-pixel-based methods in the context of gray-anchoring theory and provide theoretical analysis indicating that the two representative gray-pixel-based methods, i.e., Gray-Pixel \cite{yang2015efficient} and Grayness-Index \cite{qian2019finding}, are computationally equivalent under the Lambertian reflection model, aligning them both with the computational theory of biological color constancy. Meanwhile, a more reasonable implementation version of gray-anchoring for color constancy can be found in \cite{yang2026gray}.

\subsubsection{Gray-Pixel}
Let us begin by reviewing the formulation of gray pixel detection. The Gray-Pixel hypothesis was first proposed and verified in our previous work \cite{yang2015efficient}, which claimed that most natural images contain some gray (or at least approximately gray) pixels when being recorded under a white light source. Meanwhile, the illuminant-invariant measure is defined to identify the gray pixels from color-biased images, and hence for illuminant estimation \cite{yang2015efficient}. Specifically, under the assumption of narrow spectral response and based on the simple Lambertian model, the captured image can be normally expressed as the product of the illumination and surface reflectance, i.e.,  
\begin{equation}
\label{e1}
I^i = R^iC^i, i \in \{ r,g,b\}
\end{equation}
where $I^i$ is the captured image, while $R^i$ represents the spatial distribution of reflectance and $C^i$ denotes the spatial distribution of the illumination.

With logarithmic transform, we have
\begin{equation}
\label{e2}
log(I^i) = log(R^i)+log(C^i)
\end{equation}

According to \cite{yang2015efficient}, with the reasonable assumption that the illumination is uniform within a small local region, the illuminant-invariant measure can be defined as any local difference operator ($\Delta\{\cdot\}$) in the logarithmic space, i.e.,
\begin{equation}
\label{e3}
\Delta\{log(I^i)\} = \Delta\{log(R^i)\}+\Delta\{log(C^i)\}
\end{equation}
where $\Delta\{log(C^i)\}\approx 0$ due to the assumption of local uniform illumination, and hence, $\Delta\{log(I^i)\} = \Delta\{log(R^i)\}$ is an illuminant-invariant measure that should be independent of illumination. The $\Delta\{\cdot\}$ operator has multiple options, such as local edges or standard deviation used in the original work \cite{yang2015efficient}, local contrast in \cite{yang2018improved}, and group-based contrast \cite{cheng2024nighttime}.

Accordingly, if one pixel is in a small gray patch, the three illuminant-invariant measures in three color channels should be equal to each other for this pixel except in a flat patch, i.e., 
\begin{equation}
\label{e4}
\Delta\{log(I^r)\} = \Delta\{log(I^g)\} = \Delta\{log(I^b)\} \neq 0
\end{equation}

This is the equation proposed in our previous work\cite{yang2015efficient}, as the key criterion for gray pixel detection. In other words, the Gray-Pixel method identified the gray pixels by determining whether the differences of reflectance in each color channel are equal or not, by eliminating local illumination.

\subsubsection{Grayness-Index Under the Lambertian Model}
On the other hand, when thinking from this perspective, gray pixel detection can be achieved with different computational steps. Let us first define the luminance channel as $I^m = (I^r+I^g+I^b)$. Then, the reflectances of gray pixels in three color channels should also be equal to each other, i.e., $R^m = R^r = R^g = R^b$  and $I^m =R^m (C^r+C^g+C^b)$ for gray pixels when combining with Eq. (\ref{e1}). Thus the reflectances could be eliminated when computing the difference of cross-color channels for a gray pixel $(x,y)$. The difference values between each color channel and the luminance channel in the logarithmic space can be obtained as 
\begin{equation}
\label{e5} 
\begin{array}{l}
\varphi^i(x,y) = log(I^i)-log(I^m)  \\
 {\kern 35pt} = (log(R^i)+log(C^i))\\
 {\kern 45pt} -(log(R^m)+log(C^r+C^g+C^b))
 \end{array} 
 \end{equation}
Thus, we have $\varphi^i(x,y) = log(C^i)-log(C^r+C^g+C^b)$ for gray pixels as $R^m = R^r = R^g = R^b$, which is only related to illumination and independent of reflectance. In contrast, for the chromatic patches, $R^m = R^r = R^g = R^b$ does not hold and the $\varphi^i(x,y)$ contain both reflectance and illumination information.

\begin{figure*}[t!]
\centering
\includegraphics[width=18cm]{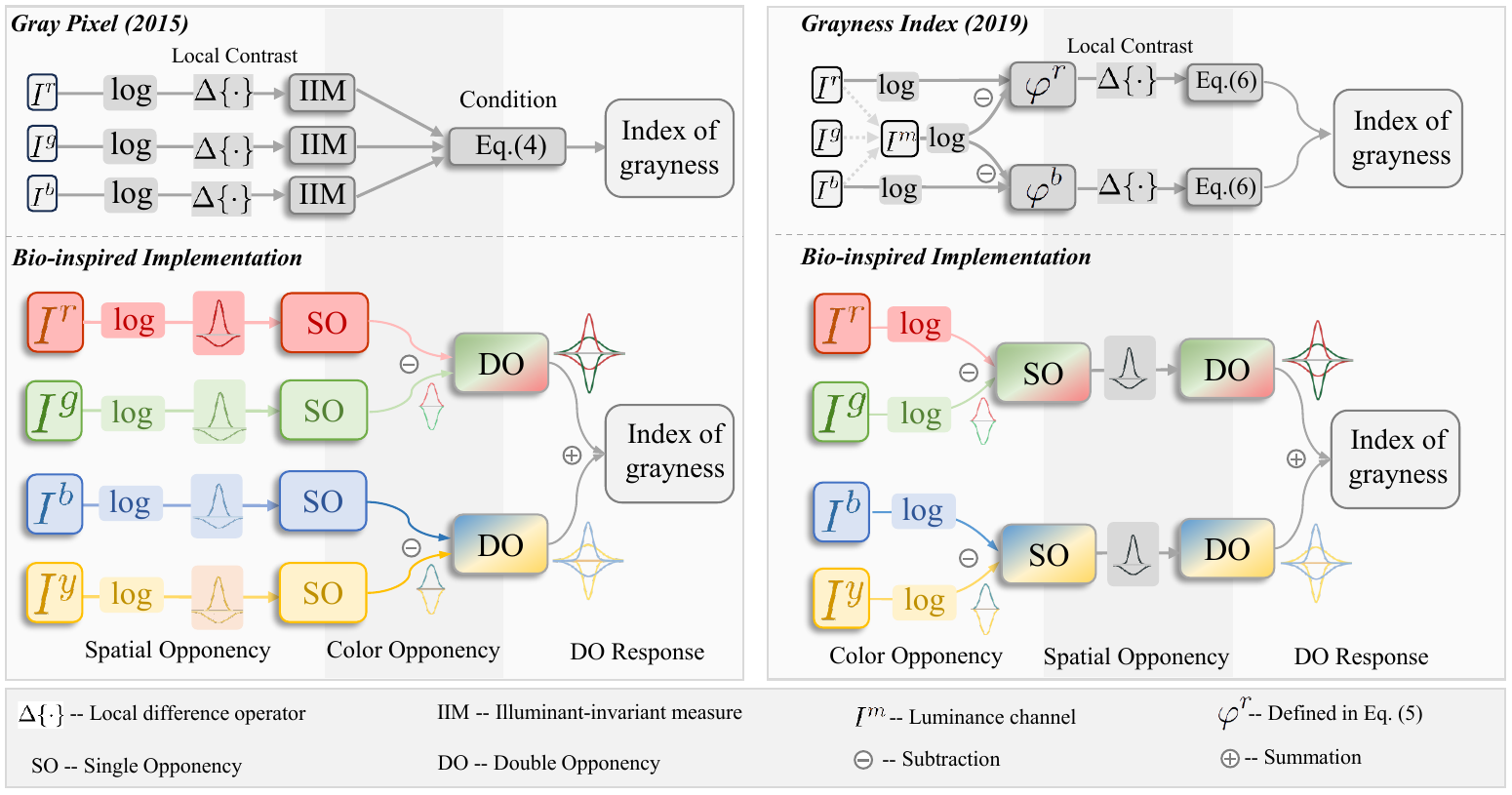}
\caption{Biological implementations of gray-pixel detection methods. \textbf{Left}: The Gray-Pixel\cite{yang2015efficient} can be implemented with double-opponent operator constructed with the first spatial single-opponency followed by color single-opponency. \textbf{Right}: The Grayness-Index \cite{qian2019finding} can be implemented with double-opponent operator constructed with the first color single-opponency followed by spatial single-opponency.}
\label{FigBioGP}
\end{figure*}

Assuming the illumination is uniform within small local patches, the local difference in $\varphi^i(x,y)$ should be zero for a gray pixel $(x,y)$. In contrast, $\varphi^i(x,y)$ still contains the difference of local reflectance and should not approach zero. Therefore, the condition for one small patch being gray is
\begin{equation}
\label{e6}
\Delta\{\varphi^r\} = \Delta\{\varphi^g\} = \Delta\{\varphi^b\} = 0
\end{equation}
Similarly, this condition only works for local non-flat regions, i.e., $\Delta\{I^i\}>\epsilon, \forall i\in \{r,g,b\}$, because a flat chromatic patch has no spatial cues. Therefore, the gray pixels can be identified by determining whether the local contrast of illumination is zero or not after eliminating the reflectance of potential gray pixels. \textit{There are some interesting things happening. This necessary condition in Eq.(\ref{e6}), according to the Lambertian model, is exactly the same as the grayness index defined by Qian et al. \cite{qian2019finding}, which is derived in the context of the dichromatic reflection model}.

In summary, we can conclude that the gray pixel detection by Qian et al. \cite{qian2019finding} and the original method by Yang et al. \cite{yang2015efficient} are computationally equivalent and can be explained in a unified way in the Lambertian model. The performance gap between the two versions on the color constancy datasets mainly arises from the specific selection of local operators and parameter settings.

\subsection{Biological Implementation of Gray Pixel Detection}
In the context of gray-anchoring theory, the two different frameworks (i.e., Gray-Pixel \cite{yang2015efficient} and Grayness-Index \cite{qian2019finding}) can be achieved by different orders of spatial and color opponency in the visual system and be linked to the bio-inspired color constancy, as suggested in \cite{yang2026gray}. In this section, we will detail the biological implementations of Gray-Pixel\cite{yang2015efficient} and Grayness-Index\cite{qian2019finding}, and hence they are classified as bio-inspired color constancy methods in the context of this study.

Figure \ref{FigBioGP} illustrates the biological implementations of these two gray-pixel-based methods. In the biological implementation of the Gray-Pixel method, the input signals are first divided into four channels ($I^r$, $I^g$,$I^b$, and $I^y =0.5(I^r+I^g)$), and then the logarithmic transform is conducted on each channel. Subsequently, when the local difference operator ($\Delta\{\cdot\}$) is defined as a center-surround operator, the illuminant-invariant measure can be equivalently computed by a spatial single-opponent receptive field in the visual system \cite{kuffler1953discharge}. That means that the responses of spatial single-opponent cells also describe the IIMs in each channel. According to Eq.(\ref{e4}), the IIMs in each color channel should be equal to each other for a gray pixel in a small gray patch. This can be equivalently implemented by a color-opponent operator by comparing the differences between IIMs in the R and G channels, or B and Y channels. Interestingly, the flowchart of spatial-opponent followed by a color-opponent operator precisely implements the function of double-opponent receptive field. In other words, we conclude that the responses of double-opponent cells indicate locations of gray pixels. Finally, double-opponent responses (R-G and B-Y double opponency) are obtained and then combined into a unified grayness index map. Therefore, the Gray-Pixel method can be well implemented with double opponency via the flowchart of first spatial single-opponency followed by color single-opponency, and hence conclude that it is biologically feasible. 

\begin{figure*}[t!]
\centering
\includegraphics[width=13cm]{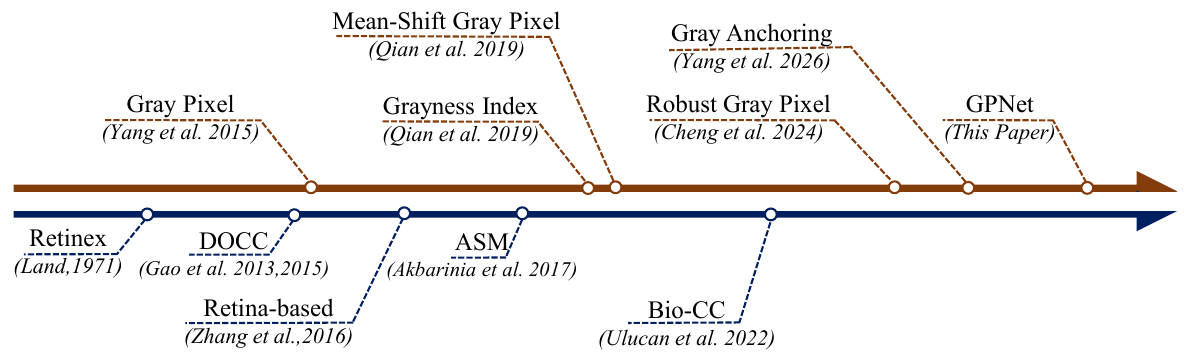}
\caption{Summarizing the development of bio-inspired color constancy methods with a timeline, including the gray-pixel family (upper part) and other methods (lower part).}
\label{FigBioCC}
\end{figure*}

On the other hand, an alternative way to implement the response of double-opponent cells is reversing the order of color and spatial single-opponency. We have no reason to eliminate this implementation considering both spatial and color single-opponent receptive fields are widely found in early vision. Specifically, color single-opponency with the center-only receptive field, i.e., Type II cells \cite{wiesel1966spatial, conway2001spatial}, could eliminate the reflectance if the pixels are gray. In other words, R-G and B-Y opponency in the logarithmic space are equivalent to the computation of difference values between each color channel and the luminance channel (Eq.(\ref{e5})). Thus, after the computation of color single-opponency, the reflectance will be eliminated and only the illumination remains in a gray patch; while partial reflectance information could be retained in a chromatic patch. According to the assumption of local uniform illumination, the degree of local smoothness of the color single-opponent responses can serve as a criterion for determining whether a patch is gray. Therefore, a local difference operator with center-surround spatial single-opponent receptive field will evaluate the smoothness of a local patch and finally exhibit the function of double-opponent receptive field. Finally, same as the Gray-Pixel method, the double-opponent responses from R-G and B-Y double opponency are combined into a unified grayness index map. The Grayness-Index\cite{qian2019finding} can also be implemented by the double-opponent cells with the flowchart of first color single-opponency followed by spatial single-opponency.

In summary, the gray-anchoring theory suggests that two computational flows with different orders of color and spatial single-opponency are computationally equivalent \cite{yang2026gray}. Moreover, they are well aligned to the computational flow of biological color constancy and can be considered as bio-inspired methods in this study. Accordingly, we summarize the typical bio-inspired color constancy methods in Fig. \ref{FigBioCC} with a timeline to show their advances in recent years. Although bio-inspired methods have evolved into a relatively systematic framework, they remain far less prevalent than deep-learning-based methods. Nevertheless, the promising potential exhibited by bio-inspired color constancy warrants greater attention.

\section{Learning to Find Gray Pixels}
\label{sec.gpnet}
In section \ref{sec.gp}, we provide basic criteria for determining whether a pixel is gray, i.e., Eq(\ref{e4}) and Eq(\ref{e6}). Unfortunately, these criteria used by Yang et al. \cite{yang2015efficient} and Qian et al. \cite{qian2019finding} are necessary conditions. That means some pixels in complex scenes (e.g., regions with cast shadows) may lead to false gray-pixel identifications. This is actually determined by the fact that illuminant estimation is an ill-posed problem. Additionally, noise commonly appear in images also influenceas the accuracy of gray pixel detection. We should not expect to completely solve this ill-posed problem, but some specific constraints are promising to improve gray pixel detection. We indeed found that different performances are reported when using different specific implementations, such as Gray-Pixel(edge), Gray-Pixel(std) \cite{yang2015efficient}, Grayness-Index \cite{qian2019finding}, and Robust Gray-Pixel \cite{cheng2024nighttime}. This motivates us to build more robust feature representations in neural scenes and further improve the performance of gray pixel detection.

In the context of prosperous feature learning techniques, we hope to leverage a neural network to learn robust features for gray pixel detection. At the same time, the model-driven and bio-inspired insights for determining gray pixel are considered as the initial constraints. We believe embedding biological constraints into a feature learning network is a reasonable trade-off that is expected to improve the performance of gray pixel detection and optimize the size of the neural network. Therefore, we build a lightweight neural network (denoted as Gray-Pixel Network, GPNet) to detect gray pixels and achieve color constancy by coupling biological insights with feature learning.

\subsection{Gray-Pixel Network}
\begin{figure*}[t]
\centering
\includegraphics[width=15cm]{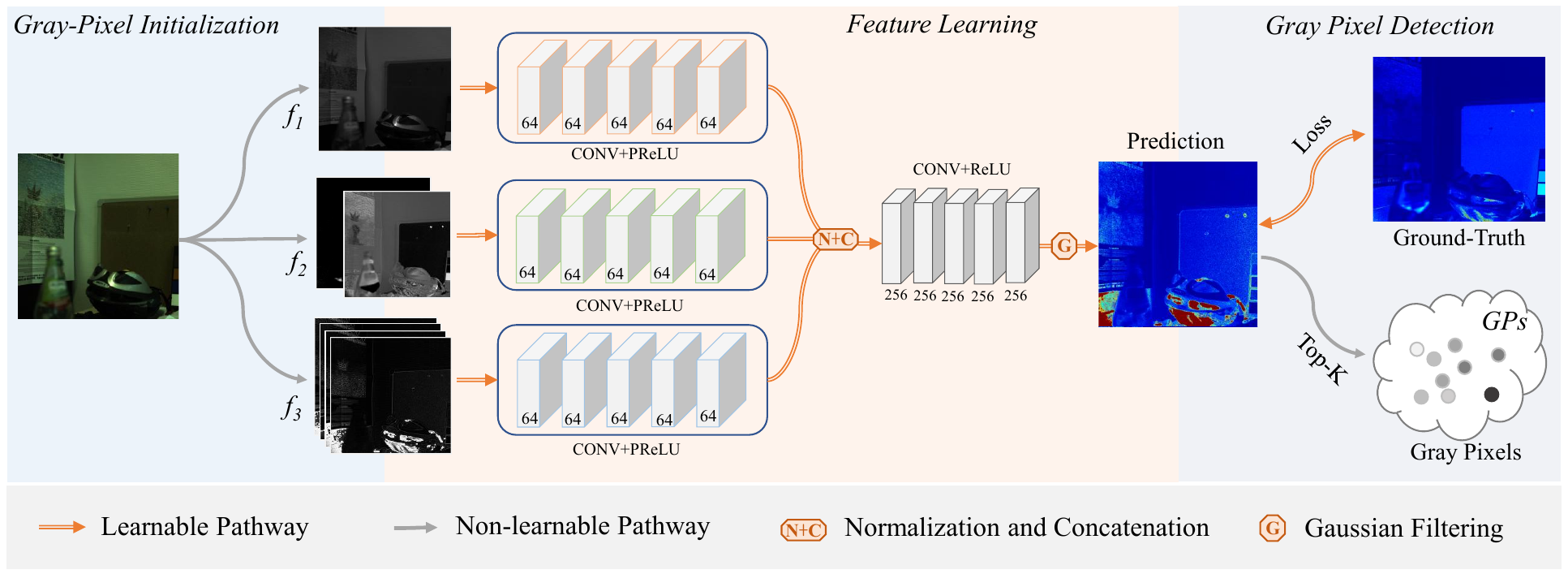}
\caption{The Gray-Pixel Network. This network begins with three main inputs driven by the initial gray-pixel constraints and then predicts the pixel-wise grayness index via lightweight convolutional modules. Finally, the Top-K pixels with the minimum gray index in the image are selected for illuminant estimation.}
\label{FigGPNet}
\end{figure*}

According to the analysis based on the Lambertian model in Section \ref{sec.gp}, eliminating the illumination or reflectance component is an important step for subsequently gray pixel detection. We take this first step as the initial constraint and follow it with a lightweight neural network to learn robust features for gray pixel prediction. The flowchart of the proposed GPNet is shown in Fig. \ref{FigGPNet}. First of all, three initial gray pixel cues are fed into a three-pathway neural network to exploit richer features. According to analysis in Section \ref{sec.gp}, gray pixel cues include the illuminant-invariant measures via eliminating local uniform illumination and the cross-channel differences for eliminating the reflectance of potential gray pixels. Additionally, the gray-scale version of the input image is also employed as a separate pathway because it was found efficient for identifying achromatic pixels by Bianco and Cusano \cite{bianco2019quasi}.

Specifically, the grayscale channel is computed as the average of the three color channels of the input image, i.e., 
\begin{equation}
\label{e7}
f_1= (I^r+I^g+I^b)
\end{equation}

For the second pathway, we compute cross-channel differences using an opponent operator to eliminate the reflectance component of potential gray pixels, consistent with color single-opponency in early vision.

\begin{equation}
\label{e8}
\begin{array}{l}
f^{rg}_2 = log(I^r)-log(I^g)   \\
f^{by}_2 = log(I^b)-log(I^y)
\end{array} 
\end{equation}
where $I^y$ denotes the yellow channel and $I^y = 0.5(I^r+I^g)$.

Finally, we eliminate local uniform illumination and obtain the illuminant-invariant measures for the four color channels via Eq.(\ref{e3}) by choosing $\Delta\{\cdot\}$ as a center-surround operator to characterize spatial single-opponency in early vision, i.e., 
\begin{equation}
\label{e9}
f^i_3 = \Delta\{log(I^i)\} =log(I^i) - F \ast log(I^i)
\end{equation}
where $F$ denotes a Gaussian filter with standard deviation as 5 pixels in all experiments and $i\in \{r,g,b,y\}$, $\ast$ denotes the convolution operator.

Subsequently, the initial gray pixel cues ($f_1$, $f_2$, and $f_3$) are fed separately into three convolutional network pathways. These three pathways share the identical network architecture except for the number of input channels, and each pathway consists of five stacked CONV+PReLU layers. It should be noted that we employ PReLU as the non-linear activation layer to retain negative values contained in the initial gray pixel cues. The outputs of the three pathways are first normalized and concatenated, then fed into another CNN module comprising five CONV+ReLU layers. Finally, a Gaussian smoothing layer is applied to the logits. The final output has the same spatial resolution as the input but with a single channel, corresponding to the predicted grayness index map.

To learn the grayness index for each pixel, we train the proposed network using a regression strategy. The loss function is defined between the predicted grayness index and its ground truth, which is generated by calculating the angular error of each pixel in the color-biased image relative to the real illuminant. However, directly using the pixel-wise loss (e.g., MSE loss) will lead to several drawbacks. First, the final illuminant estimation is based on the gray pixels, and hence the training process should firstly ensure the correct identification of potential gray pixels while relaxing constraints on chromatic pixels. Since gray pixels correspond to lower values in the target grayness index map, we design a pixel-wise weighting term as $1/\{\min(O_{xy},G_{xy})\}^2$ to force the network to focus more on potential gray pixels. Here, $O_{xy}$ denotes the predicted grayness index, and $G_{xy}$ denotes the ground-truth grayness index for the pixel at position $(x,y)$. Incorrectly assigning high grayness values to true gray pixels or low values to chromatic pixels incurs a larger penalty, as such errors would cause missing or false detection of gray pixels. Accordingly, the proposed weighted loss can be formulated as follows:

\begin{equation}
\label{e10}
L_{xy} = \frac{||{O_{xy}-G_{xy}||_2}}{\{\min(O_{xy},G_{xy})\}^2+\delta}
\end{equation}
where $\delta=0.001$ in all experiments, voiding division by zero.

Subsequently, another critical issue is the extreme imbalance between the number of near-gray pixels and chromatic pixels in natural scenes. When using the loss function defined in Eq.(\ref{e10}), feature learning will still be dominated by chromatic pixels due to their overwhelming numerical superiority. To alleviate this imbalance, the final loss is formulated based on the histogram of the grayness index map, rather than pixel-wise errors. Specifically, pixels are divided into $N$ bins according to the histogram of ground truth grayness index map, and the errors within each bin are first averaged, yielding $L_{xy}^j$ for the $j^{th}$ bin. The final loss is defined as  
\begin{equation}
\label{e11}
Loss = \sum\limits_{j = 1}^N {L_{xy}^j}
\end{equation}
Thus, regardless of the number of pixels in each bin, they are all represented by the solely mean error, thereby mitigating the imbalance issue of near-gray pixels and chromatic pixels.

In addition, several training techniques are employed to stabilize optimization. For example, the number of bins is set to $N=100$, and pixels with grayness indexes higher than 20 are assigned to the bin with the maximum index value. Furthermore,the loss is set to zero for pixels whose error is below 0.5. This prevents the network from over-focusing on pixels with negligible loss values. These hyperparameters are determined empirically through preliminary experiments and kept fixed across all subsequent experiments to avoid excessive ablation studies.

The network predicts a grayness index map for each input image, where a smaller grayness index indicates a higher probability that the pixel is a gray pixel. Accordingly, the top-K pixels with the lowest grayness index values in the image are selected as gray pixels and denoted as the set $GPs$. Finally, the illuminant is estimated from these gray pixels in the original image, i.e.,
\begin{equation}
\label{e15}
{e_i} = \frac{1}{K}\sum\limits_{(x,y) \in G{P_s}} {{I^i}(x,y), i\in\{r,g,b\} },
\end{equation}
where $K$ is the number of selected gray pixels. The final illuminant vector is $\textbf{I}_e=[e_r, e_g, e_b]^T$.

\subsection{Experimental Results}

Two widely used datasets are adopted to evaluate the proposed GPNet and validate the effectiveness of model-driven gray pixel detection for color constancy. Specifically, the employed datasets include ColorChecker\_REC (containing 568 images) and Intel\_TAU (containing 7022 images) \cite{laakom2021intel}. Notably, ColorChecker\_REC \cite{hemrit2018rehabilitating, hemrit2019providing} is used in place of the original ColorChecker dataset \cite{shi2010re}. As an updated version of ColorChecker recommended by Hemrit et al. \cite{hemrit2019providing}, ColorChecker\_REC provides a unified ground truth and standardized clipping preprocessing, ensuring consistent performance comparisons.

In previous research, the ColorChecker dataset has often been used improperly, for example with uncorrected black-level images, inconsistent ground truths, or varying clipping schemes, which can introduce inconsistencies in performance ranking during comparisons. For instance, the performance of the Gray-Pixel method \cite{yang2015efficient} reported on the ColorChecker dataset is inaccurate, due to the adoption of uncorrected black-level images. Table \ref{tab:graypixel} shows the significant performance variations of the Gray-Pixel method caused by such improper dataset usage.

The proposed GPNet is trained and evaluated on the two datasets separately. Specifically, on the ColorChecker\_REC dataset, we perform three-fold cross-validation with random data splitting. For the Intel\_TAU dataset \cite{laakom2021intel}, we follow its official ten-fold cross-validation protocol. All training is conducted on a single RTX 4090 GPU. The learning rate uses a warm-up strategy with an initial value of  $lr_0 = 0.0001$, followed by a bell-shaped scheduling that peaks at the midpoint of training and decays toward both the start and end. The number of training epochs is set to 2000 for ColorChecker\_REC and 300 for Intel\_TAU, which is roughly proportional to the number of images in each dataset.

Following the data augmentation strategy introduced in \cite{yu2020cascading}, we apply random cropping on input images, with the crop size ranging from 10\% to 100\% of the shorter side of the original image. Subsequently, the cropped patches are resized to a fixed resolution of $256\times256$ pixels, followed by a random horizontal flipping with a probability of 0.5.  Color illumination augmentation is also employed to enhance model generalization. Specifically, the illumination labels are scaled by three random factors from the range [0.6,1.4]. In contrast, we keep the input image as a linear image to  satisfy the constraints of the reflection model.

\begin{table}[t]
\caption{The inaccurate results of the Gray-Pixel method \cite{yang2015efficient} reported on the ColorChecker dataset \cite{shi2010re} due to the improper use of uncorrected black-level images. The used metric is recovery angular error.}
\label{tab:graypixel}
\centering
\begin{tabular}{l|cc|cc}
\toprule
 \textbf{ColorChecker} & \multicolumn{2}{c|}{Uncorrected} & \multicolumn{2}{c}{Correction}  \\
\midrule
Methods  & \ Median  & Mean & Median & Mean  \\
\midrule
Gray-Pixel(edge)\cite{yang2015efficient}  & 3.2 & 4.7 & 2.28 & 3.97 \\
Gray-Pixel(std) \cite{yang2015efficient} & 3.1 & 4.6 & 2.20 & 3.93 \\
\bottomrule
\end{tabular}
\end{table}

\begin{table*}[hbp]
\caption{Results on the ColorChecker\_REC dataset compared with existing methods with recovery and reproduction angular errors. Note that B.25\% and W.25\% indicate Best 25\% and Worse 25\% errors respectively.}
\label{tab:cc_rec}
\centering
\setlength{\tabcolsep}{2.2mm}{
\begin{tabular}{l|ccccc|cccccc}
\toprule
         & \multicolumn{5}{c|}{Recovery angular errors} & \multicolumn{5}{c}{Reproduction angular errors} \\
\midrule
Statistics-based Methods & Median & Mean & Trimean & B.25\% & W.25\% & Median & Mean & Trimean & B.25\% & W.25\% \\
\midrule
White-Patch  \cite{land1971lightness}   & 6.74 & 9.07 & 7.81 & 2.23 & 18.90 & 8.04 & 9.75 & 8.85 & 2.68 & 19.15\\
Gray-World  \cite{buchsbaum1980spatial} & 9.97 & 9.66 & 9.95 & 5.02 & 13.70 & 10.63 & 10.74 & 10.72 & 5.93 & 15.48\\
Shades-of-Grey \cite{finlayson2004shades}  & 6.83 & 7.25 & 6.91 & 2.29 & 12.83 & 7.55 & 8.26 & 7.75 & 2.59 & 15.02\\
General Gray-World \cite{barnard2002comparison} & 5.95 & 6.60 & 6.11 & 1.95 & 12.44 & 6.66 & 7.57 & 6.97 & 2.33 & 14.14\\
Gray-Edge-$1^{st}$  \cite{van2007edge} & 3.09 & 4.02 & 3.30 & 0.95 & 8.74 & 3.74 & 4.96 & 4.05 & 1.09 & 10.91\\
Gray-Edge-$2^{nd}$  \cite{van2007edge} & 3.57 & 4.38 & 3.79 & 1.51 & 8.48 & 4.54 & 5.45 & 4.78 & 1.90 & 10.49\\
LSRS \cite{gao2014efficient}  & 2.88 & 3.80 & 3.14 & 1.33 & 7.71 & 3.75 & 4.66 & 3.97 & 1.56 & 9.36 \\

\midrule
Learning-based Methods & Median  & Mean & Trimean & B.25\% & W.25\% & Median & Mean & Trimean & B.25\% & W.25\% \\
\midrule
Pixel-based Gamut \cite{gijsenij2010generalized} & 4.41 & 6.01 & 4.88 & 1.71 & 12.90 & 5.20 & 6.88 & 5.67 & 2.02 &  14.48\\
Edge-based Gamut \cite{gijsenij2010generalized} & 3.27 & 5.48 &  3.89 & 0.68 & 13.78 & 4.56 & 6.86 & 5.19 & 0.85 &  16.75 \\
Intersection-Based Gamut \cite{gijsenij2010generalized} & 4.41 & 6.00 & 4.86 & 1.71 & 12.78 & 5.20 & 6.83 & 5.63 &  2.02 & 14.32\\
C3AE \cite{laakom2019color} & 1.9 & 2.1 & 2.0 & 0.8 & 4.0 & -- & -- & -- & -- & --\\
BoFC \cite{laakom2020bag} & 1.2 & 2.0 & 1.4 & 0.3 & 4.8 & -- & -- & -- & -- & -- \\
C4 \cite{yu2020cascading} & 0.98 & 1.43 & 1.05 & 0.29 & 3.38 & -- & -- & -- & -- & --\\
C5 \cite{afifi2021cross} & 1.40 & 2.20 & 1.57 & 0.37 & 5.41 & -- & -- & -- & -- & -- \\
\midrule
Bio-inspired Methods & Median  & Mean & Trimean & B.25\% & W.25\% & Median & Mean & Trimean & B.25\% & W.25\% \\
\midrule
Bio-CC \cite{ulucan2022bio}  & 3.30 & 4.40 & -- & 0.86 &  9.84 & -- &  -- &   --  &  -- &  --\\
DOCC \cite{gao2015color} & 2.78 & 4.76 & 3.11 & 0.63 & 12.46 & 3.58 & 5.89 & 4.07 & 0.76 & 15.14\\
Gray-Pixel(edge) \cite{yang2015efficient} & 2.43 & 4.58 & 2.83 & 0.52 & 12.34 & 3.16 & 5.65 & 3.79 & 0.64 & 14.82 \\
Gray-Pixel(std) \cite{yang2015efficient} & 2.55 & 4.57 & 2.89 & 0.53 & 12.23 & 3.31 & 5.61 & 3.85 & 0.66 & 14.61 \\
Grayness-Index \cite{qian2019finding} & 1.91 & 3.20 & 2.21 & 0.44 & 8.01 & 2.48 & 4.15 & 2.93 & 0.56 & 10.43\\
\textbf{GPNet (K=5000)} & 1.41 & 2.31 & 1.64 & 0.36 & 5.65 & 1.80 & 3.00 & 2.13 & 0.43 & 7.44 \\
\bottomrule
\end{tabular}}
\end{table*}

\begin{figure*}[htp]
\centering
\includegraphics[width=17cm]{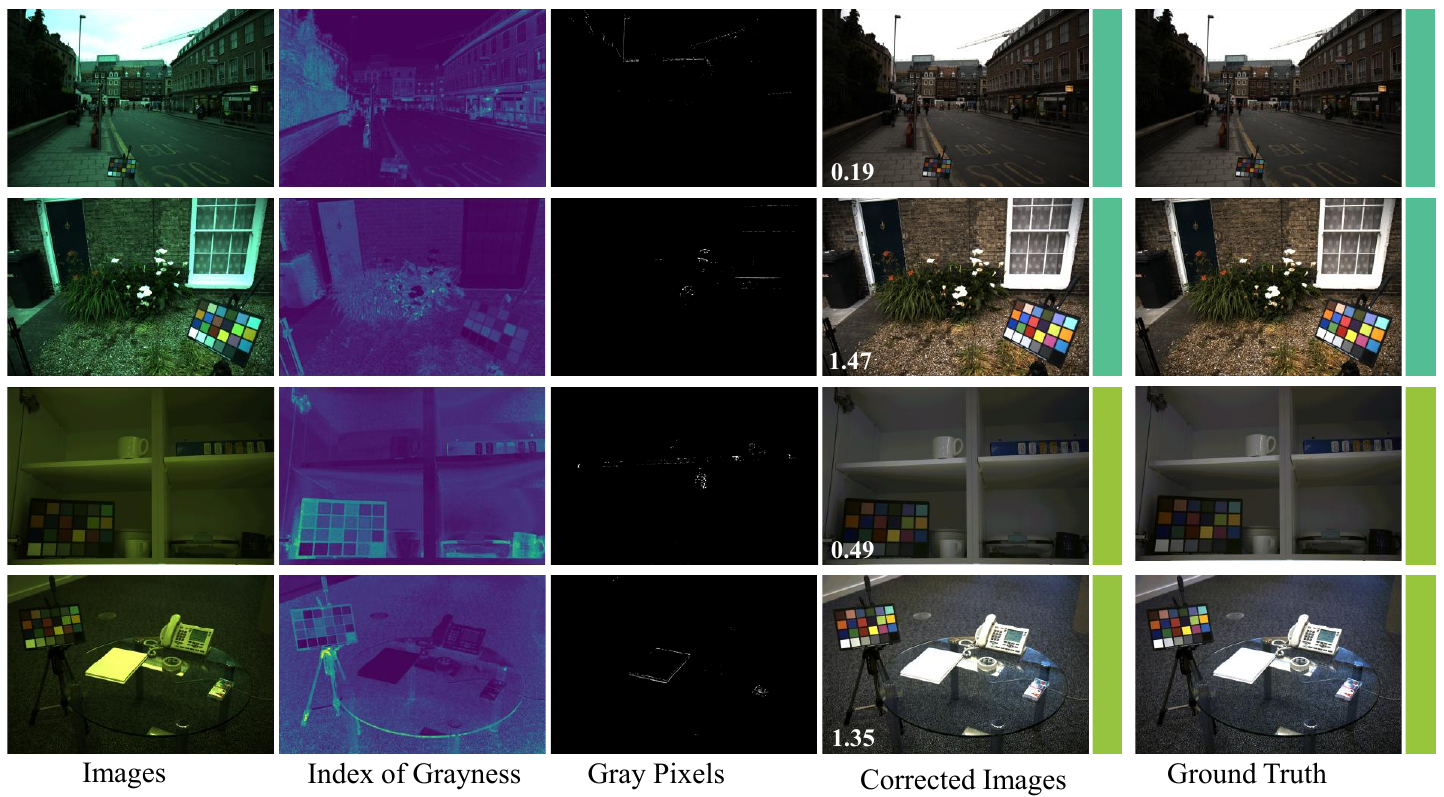}
\caption{Visualization results of the proposed GPNet. \textbf{From left to right}: input images, predicted grayness index, selected gray pixels (top 5000 pixels), corrected images and estimated illuminants, the ground truth images and illuminants. Macbeth Color Checker is always masked when choosing gray pixels. Numbers in results indicate recovery angle errors.}
\label{FigResults}
\end{figure*}

Table \ref{tab:cc_rec} presents the experimental results on the ColorChecker\_REC dataset, comparing the proposed method with existing methods in terms of both recovery angular error \cite{hordley2006reevaluation, gijsenij2009perceptual} and reproduction angular error \cite{finlayson2016reproduction}. For gray-pixel-based methods (i.e., Gray-Pixel and Grayness-Index), the proportion of gray pixels used for illuminant estimation is uniformly set to 0.1\% in all experiments. Meanwhile, our GPNet selects 5000 gray pixels via the Top-K operator (K=5000), which corresponds to roughly 0.1\% of all pixels.

Notably, many deep learning-based methods are not included in Table \ref{tab:cc_rec}, since most of them are evaluated on the  ColorChecker dataset \cite{shi2010re} rather than ColorChecker\_REC. Nevertheless, C3AE \cite{laakom2019color} and BoFC \cite{laakom2020bag} are included for comparison, as they explicitly report performance on the ColorChecker\_REC dataset. Furthermore, we also add two state-of-the-art methods, C4 \cite{yu2020cascading} and C5 \cite{afifi2021cross}, which we retrained on the ColorChecker\_REC dataset using their publicly available source code.

Figure \ref{FigResults} shows the visualization results of the proposed GPNet for gray pixel detection and color constancy. The GPNet produces reasonable grayness index maps for both outdoor and indoor scenes. In the grayness index, darker blue indicates a higher degree of grayness. Meanwhile, the selected gray pixels typically lie on achromatic surfaces and tend to be located at high-contrast edges.

Overall, deep learning-based methods consistently outperform their statistical counterparts. Furthermore, bio-inspired methods generally outperform classical statistical or gamut-based techniques \footnote{Results on the ColorChecker\_REC dataset are directly from http://www.colorconstancy.com/.} while achieving performance comparable to deep-learning methods considered. Specifically, C4 \cite{yu2020cascading} achieves the best performance by employing a relatively large cascade-structured network. In contrast, the proposed GPNet uses compact convolutional modules and primarily leverages local initial features constrained by gray pixels. As a result, GPNet effectively integrates model-driven principles with data-driven feature learning, further improving the performance of the gray-pixel algorithm family. This demonstrates that illuminant estimation via robust gray pixel detection remains a promising and effective direction for future research.

\begin{table*}[t]
\caption{Results on the Intel\_TAU dataset compared with existing methods with recovery and reproduction angular errors. Note that B.25\% and W.25\% indicate Best 25\% and Worse 25\% errors respectively.}
\label{tab:tau}
\centering
\setlength{\tabcolsep}{2.2mm}{
\begin{tabular}{l|ccccc|cccccc}
\toprule
         & \multicolumn{5}{c|}{Recovery angular errors} & \multicolumn{5}{c}{Reproduction angular errors} \\
\midrule
 Methods & Median & Mean & Trimean & B.25\% & W.25\% & Median & Mean & Trimean & B.25\% & W.25\% \\
\midrule
Gray-World \cite{buchsbaum1980spatial}       & 3.9 & 4.9 & 4.1 & 1.0 & 10.5 & 4.9 & 6.1 & 5.2 & 1.2 & 13.0 \\
White-Patch \cite{land1971lightness}         & 9.1 & 9.4 & 9.2 & 1.4 & 17.6 & 9.5 & 10.0 & 9.8 & 1.8 & 19.2 \\
Gray-Edge-$1^{st}$  \cite{van2007edge}       & 4.0 & 5.9 & 4.6 & 1.0 & 13.8 & 4.9 & 6.8 & 5.5 & 1.2 & 13.5 \\
Gray-Edge-$2^{nd}$  \cite{van2007edge}       & 3.9 & 6.0 & 4.8 & 1.0 & 14.0 & 4.9 & 6.9 & 5.6 & 1.2 & 15.7 \\
Shades-of-Grey \cite{finlayson2004shades}    & 3.8 & 5.2 & 4.3 & 0.9 & 11.9 & 4.7 & 6.3 & 5.1 & 1.1 & 13.9 \\
PCA-based \cite{cheng2014illuminant} & 3.2 & 4.5 & 3.5 & 0.7 & 10.6 & 4.0 & 5.5 & 4.4 & 0.9 & 12.7 \\
Weighted Gray-Edge \cite{gijsenij2009physics}        & 3.7 & 6.1 & 4.6 & 0.8 & 15.1 & 4.5 & 6.9 & 5.4 & 1.1 & 16.5 \\
Color Tiger \cite{banic2017unsupervised}         & 2.6 & 4.2 & 3.2 & 1.0 & 9.9 & 3.3 & 5.3 & 4.1 & 1.1 & 12.7 \\
PCC\_Q2 \cite{d2013probabilistic}               & 2.4 & 3.9 & 2.8 & 0.6 & 9.6 & 3.5 & 5.1 & 4.0 & 0.7 & 11.9 \\
FFCC \cite{barron2017fast}         & 1.6 & 2.4 & 1.8 & 0.4 & 5.6 & 2.1 & 3.0 & 2.3 & 0.5 & 7.1 \\
CNN(Bianco) \cite{bianco2015color}             & 2.6 & 3.5 & 2.8 & 0.9 & 7.4 & 3.4 & 4.4 & 3.6 & 1.1 & 9.4 \\
C3AE \cite{laakom2019color}                 & 2.7 & 3.4 & 2.8 & 0.9 & 7.0 & 3.3 & 3.9 & 3.5 & 1.1 & 8.8 \\
BoCF \cite{laakom2020bag}                 & 1.9 & 2.4 & 2.0 & 0.7 & 5.1 & 2.3 & 3.0 & 2.5 & 0.8 & 6.5 \\
FC4 (VGG16) \cite{hu2017fc4}           & 1.7 & 2.2 & 1.8 & 0.6 & 4.7 & 2.2 & 2.9 & 2.3 & 0.7 & 6.1 \\
\midrule
Bio-inspired Methods & Median  & Mean & Trimean & B.25\% & W.25\% & Median & Mean & Trimean & B.25\% & W.25\% \\
\midrule
Bio-CC \cite{ulucan2022bio}  & 3.05 & 4.14 & -- & 0.76 & 9.42 & -- &  -- &   --  &  -- &  --\\
DOCC \cite{gao2015color} & 2.64 & 4.43 & 3.16 & 0.63 & 11.15 & 3.25 & 5.10 & 3.81 & 0.80 & 12.35 \\
Grayness-Index \cite{qian2019finding}  & 2.46 & 3.90 & 2.83 & 0.56 & 9.60 & 3.16 & 4.99 & 3.66 & 0.70 & 12.21 \\
Gray-Pixel(edge) \cite{yang2015efficient} & 2.02 & 3.08 & 2.23 & 0.53 & 7.47 & 2.57 & 3.94 & 2.86 & 0.65 & 9.57 \\
Gray-Pixel(std) \cite{yang2015efficient} & 1.99 & 2.99 & 2.19 & 0.52 & 7.16 & 2.55 & 3.83 & 2.82 & 0.65 & 9.21 \\
\textbf{GPNet (K=5000)}                & 1.74 & 2.66 & 1.94 & 0.43 & 6.44 & 2.28 & 3.42 & 2.50 & 0.54 & 8.36 \\
\bottomrule
\end{tabular}}
\end{table*}

Additional experimental results on the Intel\_TAU dataset are presented in Table \ref{tab:tau}. We draw consistent conclusions with those obtained on the ColorChecker\_REC dataset, i.e., the proposed GPNet yields performance comparable to state-of-the-art deep learning methods.

\subsubsection{Ablation Study}
Several specific settings related to the loss function design (see Eqs.~\ref{e10} and \ref{e11}) in Section~\ref{sec.gpnet} are kept fixed throughout all experiments. These parameters were determined empirically via preliminary experiments and may not be strictly optimal; nevertheless, they yield acceptable performance. We thus maintain these fixed settings to balance model effectiveness and avoid excessive ablation studies.

In this section, we perform two main ablation studies on the ColorChecker\_REC dataset. First, we check the contribution of the initial gray-pixel cues. Specifically, we replace all three initial gray-pixel cues with input images, while using the same network architecture to predict gray pixels (denoted as \textit{GPNet-w/o-GP(K=5000)}). In this configuration, the network degenerates into a standard CNN that directly estimates the grayness index map from color-biased inputs. Meanwhile, we retain the three-pathway structure to ensure the number of parameters remains comparable to that of GPNet. Experimental results reported in Table \ref{tab:as} show that, when removing the constraints of initial gray-pixel cues, performance drops to 1.97 (median) and 3.01 (mean) in terms of recovery angular error on the ColorChecker\_REC dataset.

In addition, we further investigate the influence of selecting a different number of gray pixels for illuminant estimation, i.e, the different values of $K$ setting in the top-K operator. As shown in Table \ref{tab:as}, varying K affects the performance to some extent, but the proposed model remains relatively robust overall. For instance, the median recovery angular errors stays at or below 1.5 when $K$ varies from 1000 to 10000.

\begin{table}[t]
\caption{Ablation studies on the ColorChecker\_REC dataset. The used metric is recovery angular error.}
\label{tab:as}
\centering
\begin{tabular}{l|cc}
\toprule
\midrule
Methods  & Median & Mean \\
\midrule
GPNet-w/o-GP(K=5000) & 1.97 & 3.01 \\
GPNet(K=5000) & 1.41 & 2.31  \\
\midrule
GPNet(K=1000) & 1.52 & 2.32 \\
GPNet(K=3000) & 1.45 & 2.30 \\
GPNet(K=5000) & 1.41 & 2.31 \\
GPNet(K=7000) & 1.44 & 2.34 \\
GPNet(K=9000) & 1.48 & 2.36 \\
\bottomrule
\end{tabular}
\end{table}

\section{Conclusion and Discussion}
This study proposes a comprehensive technical route that integrates biological mechanisms, computational theory, and algorithmic implementation for bio-inspired color constancy. We verified that the gray anchor detection is a potential strategy for illuminant estimation and color constancy in early vision. Finally, typical gray-pixel detection methods are unified theoretically and angling to gray anchoring theory of the biological visual system. Learning-based gray-pixel detection is also proposed to explore the potential of bio-inspired color constancy in sufficient experiments.

Gray pixel detection has also shown promising potential for visual enhancement in more complex scenes. For example, in haze conditions, Zhang et al. \cite{zhang2021haze} pointed out based on the atmospheric scattering model that when the medium transmittance is sufficiently high, the captured image is dominated by direct attenuation rather than air light. On this basis, gray pixels in high-transmittance regions can be detected and utilized for color restoration. We believe that gray pixel detection also has great potential in other tasks relying on the atmospheric scattering model, such as underwater image enhancement. In addition, the illuminant-invariant representation has also been verified to benefit object classification under varying illumination, even when applied to non-linear images \cite{funt2022laplacian}.

\section*{Acknowledgments}
This study was supported by the STI2030-Major Projects (2022ZD0204600) and the National Natural Science Foundation of China (T2541042, 62476050, 32571280). 

\bibliographystyle{IEEEtran}
\bibliography{IEEEabrv,Refs}



\begin{IEEEbiography}[{\includegraphics[width=1.2in,height=1.25in,clip,keepaspectratio]{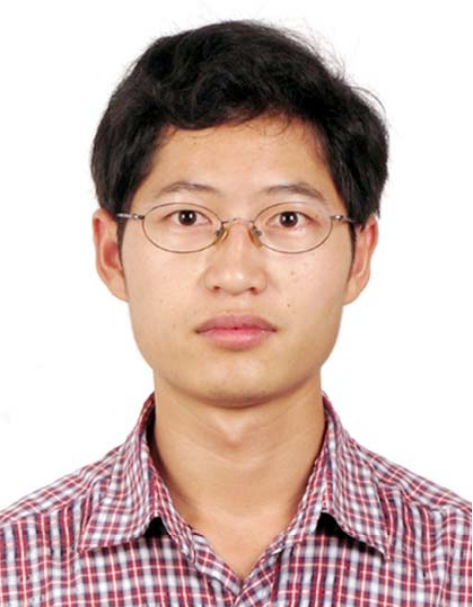}}]{Kai-Fu Yang}(Member, IEEE) 
received the Ph.D. degree in biomedical engineering from the University of Electronic Science and Technology of China (UESTC), Chengdu, China, in 2016. He is currently an Associate Research Professor with the MOE Key Laboratory for Neuroinformation, School of Life Science and Technology, UESTC. His research interests include cognitive computing and bio-inspired computer vision.
\end{IEEEbiography}

\vspace{11pt}

\begin{IEEEbiography}[{\includegraphics[width=1.2in,height=1.25in,clip,keepaspectratio]{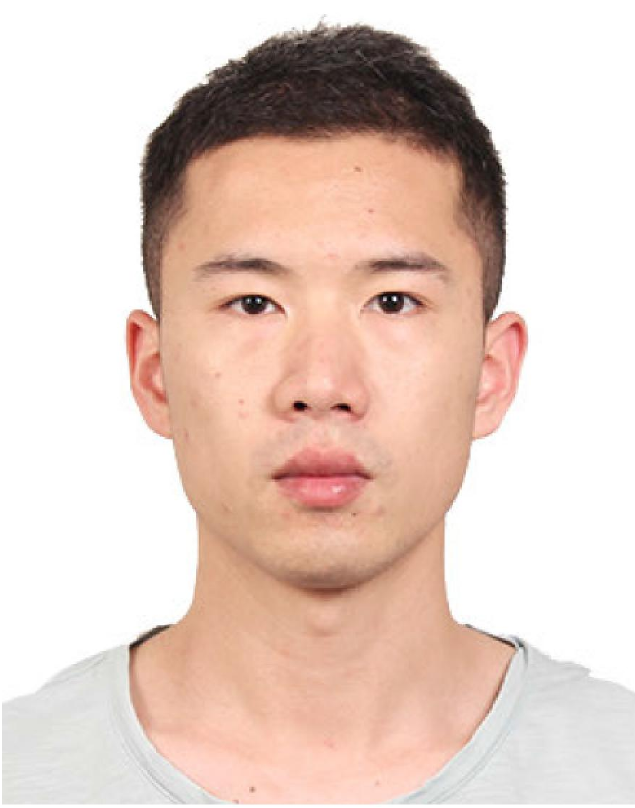}}]{Fu-Ya Luo}
received the Ph.D. degree in biomedical engineering from the University of Electronic Science and Technology of China (UESTC) in 2023. He is currently a Lecturer with the School of Electronic Engineering and Automation, Guilin University of Electronic Technology. His research interests include scene understanding, brain-inspired computer vision, weakly supervised learning, and image-to-image translation.
\end{IEEEbiography}

\vspace{11pt}

\begin{IEEEbiography}[{\includegraphics[width=1.2in,height=1.25in,clip,keepaspectratio]{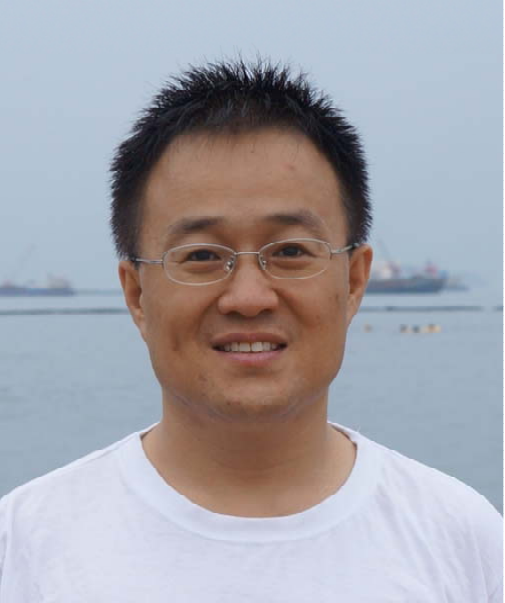}}]{Yong-Jie Li}
(M'14-SM'17) received the Ph.D. degree in biomedical engineering from the University of Electronic Science and Technology of China (UESTC), Chengdu, China, in 2004. He is currently a Professor with the MOE Key Lab for Neuroinformation, School of Life Science and Technology, UESTC. His research interests include visual mechanism modeling, image processing, and intelligent computation.
\end{IEEEbiography}

\vfill
\end{document}